\documentclass[conference]{IEEEtran}
\usepackage{cite}
\usepackage{amsmath,amssymb,amsfonts}
\usepackage{algorithmic}
\usepackage{graphicx}
\usepackage{textcomp}
\usepackage{xcolor}
\usepackage{amsmath,amssymb,amsfonts}
\usepackage{graphicx}
\usepackage{textcomp}
\usepackage{xcolor}
\usepackage{booktabs} 
\usepackage{multirow}
\usepackage{hyperref}
\usepackage{tabularx}
\usepackage[switch]{lineno}  
\usepackage{rotating}
\usepackage{siunitx}
\usepackage{comment}
\usepackage{enumitem}
\usepackage{mathtools}
\usepackage{amssymb}
\usepackage{algorithm}

\hypersetup{
    colorlinks=true,
    urlcolor=cyan,
    }

\def\BibTeX{{\rm B\kern-.05em{\sc i\kern-.025em b}\kern-.08em
    T\kern-.1667em\lower.7ex\hbox{E}\kern-.125emX}}

\title{Counterfactual Gradients-based Quantification of Prediction Trust in Neural Networks}

\author{\IEEEauthorblockN{Mohit Prabhushankar and Ghassan AlRegib}
\IEEEauthorblockA{\textit{OLIVES at the Center for Signal and Information Processing (CSIP)} \\ 
\textit{School of Electrical and Computer Engineering} \\
\textit{Georgia Institute of Technology}\\
Atlanta, GA, USA \\
\{mohit.p, alregib\}@gatech.edu}
}

\begin{document}

\onecolumn 
\begin{description}[leftmargin=2cm,style=multiline]

\item[\textbf{Citation}]{M. Prabhushankar and G. AlRegib, "Counterfactual Gradients-based Quantification of Prediction Trust in Neural Networks", In 2024 IEEE 7th International Conference on Multimedia Information Processing and Retrieval (MIPR), San Jose, CA, Aug. 7-9, 2024 (Invited Paper)}

\item[\textbf{Review}]{Data of Invitation : 31 Jan 2024 \\ Date of Invitation Acceptance: 1 Feb 2024 \\ Date of Initial Submission : 27 April 2024 \\ Date of Accept: 20 May 2024}

\item[\textbf{Codes}]{\url{https://github.com/olivesgatech/GradTrust}}

\item[\textbf{Copyright}]{\textcopyright 2024 IEEE. Personal use of this material is permitted. Permission from IEEE must be obtained for all other uses, in any current or future media, including reprinting/republishing this material for advertising or promotional purposes,
creating new collective works, for resale or redistribution to servers or lists, or reuse of any copyrighted component
of this work in other works. }

\item[\textbf{Contact}]{\href{mailto:mohit.p@gatech.edu}{mohit.p@gatech.edu}  OR \href{mailto:alregib@gatech.edu}{alregib@gatech.edu}\\ \url{https://alregib.ece.gatech.edu/} \\ }
\end{description}

\thispagestyle{empty}
\newpage
\clearpage
\setcounter{page}{1}

\twocolumn

\maketitle

\begin{abstract}
The widespread adoption of deep neural networks in machine learning calls for an objective quantification of esoteric trust. In this paper we propose \texttt{GradTrust}, a classification trust measure for large-scale neural networks at inference. The proposed method utilizes variance of counterfactual gradients, i.e. the required changes in the network parameters if the label were different. We show that \texttt{GradTrust} is superior to existing techniques for detecting misprediction rates on $50000$ images from ImageNet validation dataset. Depending on the network, \texttt{GradTrust} detects images where either the ground truth is incorrect or ambiguous, or the classes are co-occurring. We extend \texttt{GradTrust} to Video Action Recognition on Kinetics-400 dataset. We showcase results on $14$ architectures pretrained on ImageNet and $5$ architectures pretrained on Kinetics-400. We observe the following: (i) simple methodologies like negative log likelihood and margin classifiers outperform state-of-the-art uncertainty and out-of-distribution detection techniques for misprediction rates, and (ii) the proposed \texttt{GradTrust} is in the Top-2 performing methods on $37$ of the considered $38$ experimental modalities. The code is available at: \url{https://github.com/olivesgatech/GradTrust}
\end{abstract}


\section{Introduction}
\label{sec:Intro}
Machine Learning (ML) provides data driven tools that enable the construction and adoption of Artificial Intelligence (AI) systems. To garner the trust of humans who use these AI systems, the underlying ML models must lend themselves to certain attributes that fall under the umbrella terminology of trustworthiness. The study of these attributes is complex since both trust and trustworthiness are functions of the scientific communities that consider them. For instance, in the field of autonomous vehicles, the algorithmic trust in the vehicle's perception module is different from the moral trust that is placed upon it, which is again different from the governmental policy trust that the vehicle abides by. Characterizing and classifying the attributes of trustworthiness is currently an active research field. The authors in~\cite{toreini2020relationship} identify four attributes of trustworthiness that impact the trust in AI systems. These include fairness, explainability, auditability and safety of ML models. Note that these attributes are application areas within ML. The author in~\cite{varshney2022trustworthy} pedagogically describes all applications in ML that validate and satisfy the trustworthiness attributes of performance, reliability, human interaction, and aligned purpose. These applications, in the field of computer vision, include explainability~\cite{alregib2022explanatory}, out-of-distribution detection~\cite{lee2023probing, huang2021importance}, adversarial detection~\cite{lee2022gradient}, causal analysis~\cite{prabhushankar2021extracting}, noise robustness~\cite{prabhushankar2022introspective}, anomaly detection~\cite{kwon2020backpropagated}, differential privacy~\cite{abadi2016deep}, and visual uncertainty quantification~\cite{kendall2017uncertainties} among others. Hence, trustworthiness attributes are closely linked to the underlying ML applications with each application satisfying certain attributes. These trustworthiness attributes enhance the system-level esoteric trust placed in ML models. 

In this paper, we tackle trust from an individual test-sample level by providing trust quantification for every predictive testing sample. A widely used but often incorrect quantification of prediction trust is the softmax probability of the output. The authors in~\cite{gal2016dropout} show that softmax probability of a testing sample is overestimated even when the sample is far from the training data domain. Early works on prediction trust provide an algorithmic definition of trust. This definition is based on subjective logic~\cite{josang2006trust}, where trust is a quantification that is dependent on belief, uncertainty, and apriori probability of new data. On simple classifiers and small-scale neural networks, the authors in~\cite{jiang2018trust} describe trust for the application of classification as agreement between the classifier and a modified nearest-neighbor classifier on the testing example. However, the trust score in both~\cite{cheng2020there} and~\cite{jiang2018trust} is not transferable to modern large-scale neural networks. The subjective-logic based parametric assumptions in~\cite{cheng2020there} and creating nearest neighbor classifiers in~\cite{jiang2018trust} at test-time are infeasible.

\begin{figure*}
  \includegraphics[width=\textwidth]{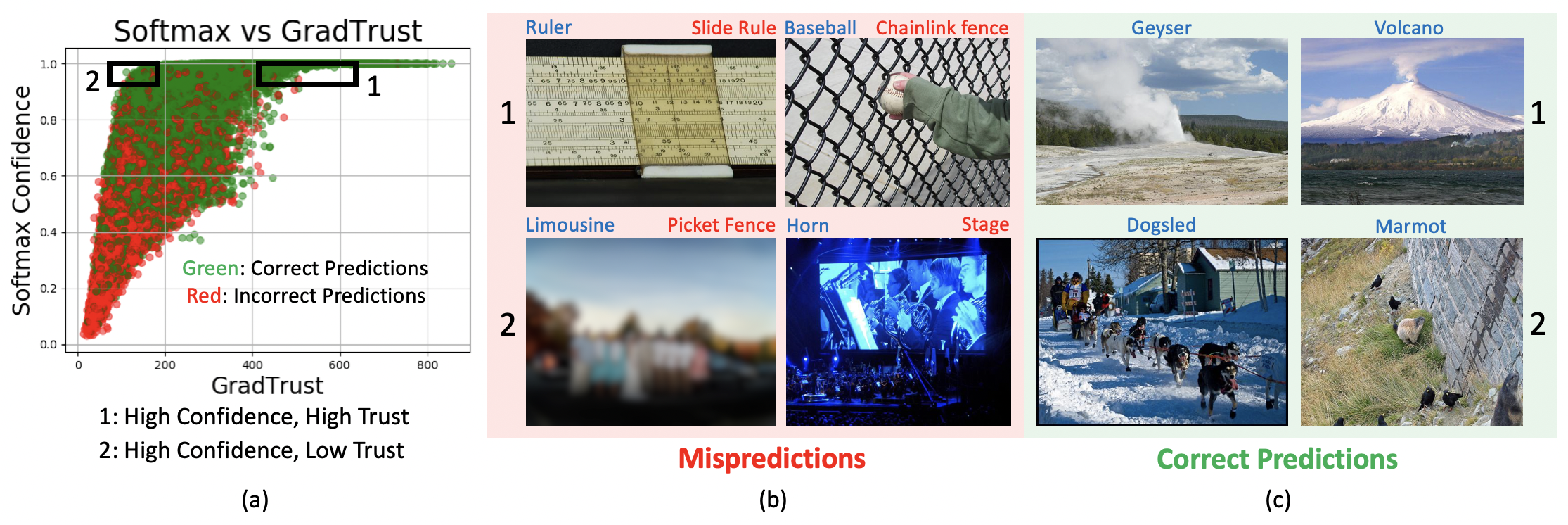}
  \caption{Scatter plot between the proposed \texttt{GradTrust} on \texttt{x-axis} and softmax confidence on \texttt{y-axis} on ImageNet validation dataset using ResNet-18. Green points indicate correctly classified data and red indicates misclassified data. Representative misclassified and correctly images in the numbered boxes are displayed alongside the scatterplot, with their predictions (in red) and labels (in blue).}
  \label{fig:Concept}
\end{figure*}

In this paper we propose \texttt{GradTrust}, an inferential quantification of trust for the application of classification. We leverage new proposals of utilizing gradients as features~\cite{lee2023probing, prabhushankar2022introspective, huang2021importance} to provide \texttt{GradTrust} scores for every test sample. The effectiveness and utility of \texttt{GradTrust} is depicted in Fig.~\ref{fig:Concept}. All images in the validation set of ImageNet dataset~\cite{deng2009imagenet} are passed through a pretrained ResNet-18~\cite{he2016deep} architecture and each image's softmax confidence is plotted against it's \texttt{GradTrust} score in Fig.~\ref{fig:Concept}a. To visually analyze \texttt{GradTrust}, we select ResNet-18's mispredictions (reds in scatter plot) and correct predictions (greens in scatter plot) from within boxes 1 and 2 in the scatter plot and display representative images in Figs.~\ref{fig:Concept}b and~\ref{fig:Concept}c respectively. Row 1 consists of images chosen from box 1 and row 2 consists of images chosen from box 2. We first analyze the four mispredicted images in Fig.~\ref{fig:Concept}b. The true label is shown above each image in blue while it's prediction is shown in red. Box 1 encompasses samples with high softmax and \texttt{GradTrust} scores. It can be visually observed that in the case of the ruler, the true label is incorrect and the network's trust in it's prediction is not misplaced. Similarly, both baseball and chainlink fence are present in the second image. In box 2, the displayed images have high softmax confidence but low \texttt{GradTrust}. Softmax places a high confidence in the image that is incorrectly labeled as picket fence. \texttt{GradTrust}, on the other hand, provides a low trust score, on account of he limousine in the background (this image is blurred to protect the privacy of the people in the photograph). Similarly, both horn and stage are present in the second image. While softmax is very confident, \texttt{GradTrust} is less inclined to trust the network's prediction. In Section~\ref{sec:Experiments}, we evaluate trust quantification as misprediction rate on $14$ deep architectures. 
\section{Related Work}
\label{sec:Related_Work}
\noindent\textbf{Softmax Confidence} Classification models are trained to optimize towards predicting the correct class and not the correct probabilities. Hence, at inference, an additional computation is required for confidence estimation, with softmax being the most widely used measure. However, the authors in~\cite{gal2016dropout} show that softmax probabilities do not estimate either uncertainty or confidence. Fig.~\ref{fig:Concept}b is a visual representation of softmax's misplaced confidence. 

\noindent\textbf{Uncertainty Quantification} Uncertainty quantification deals with assigning probabilities regarding decisions made under some unknown states of the system. A number of techniques including~\cite{gal2016dropout, van2020uncertainty, 10053381, zhou2022ramifications, lee2022gradient, lee2023probing} quantify uncertainty as a probability score. However, a majority of these methods require additional modeling~\cite{van2020uncertainty}, data~\cite{10053381}, and labels~\cite{zhou2022ramifications}. In this paper, we empirically analyze two methods of uncertainty against trust, Monte-Carlo Dropout~\cite{gal2016dropout} and GradPurview~\cite{lee2023probing}. Uncertainty is closely related to trust analysis. The authors in~\cite{cheng2020there} model uncertainty as one component within trust along with prior belief. Specifically, they show that uncertainty is not always malicious. Rather, networks trained with uncertain data positively impact belief thereby creating trustworthy models. We empirically verify that well-trained models provide good trust and uncertainty quantification scores for the application of misprediction detection in Section~\ref{sec:Experiments}.

\noindent\textbf{Out-of-Distribution (OOD) Detectors} OOD detection is a form of indirect trust quantification where the application goal is to construct a quantifier that measures if a given test sample is in-distribution or out-of-distribution as compared to the training data~\cite{kwon2020backpropagated}. While not necessarily defining trust, methods like~\cite{kwon2020backpropagated, kwon2020novelty} propose algorithms that exhibit trustworthy attributes. However, the ever evolving landscape of indirect trust quantification applications create methods that are application-specific. For instance, we show in Section~\ref{sec:Experiments} that methods like~\cite{lee2023probing} and~\cite{huang2021importance} that excel in Out-of-Distribution detection suffer in misprediction detection experiments.

\begin{figure*}[t!]
    \centering
    \includegraphics[width=\linewidth]{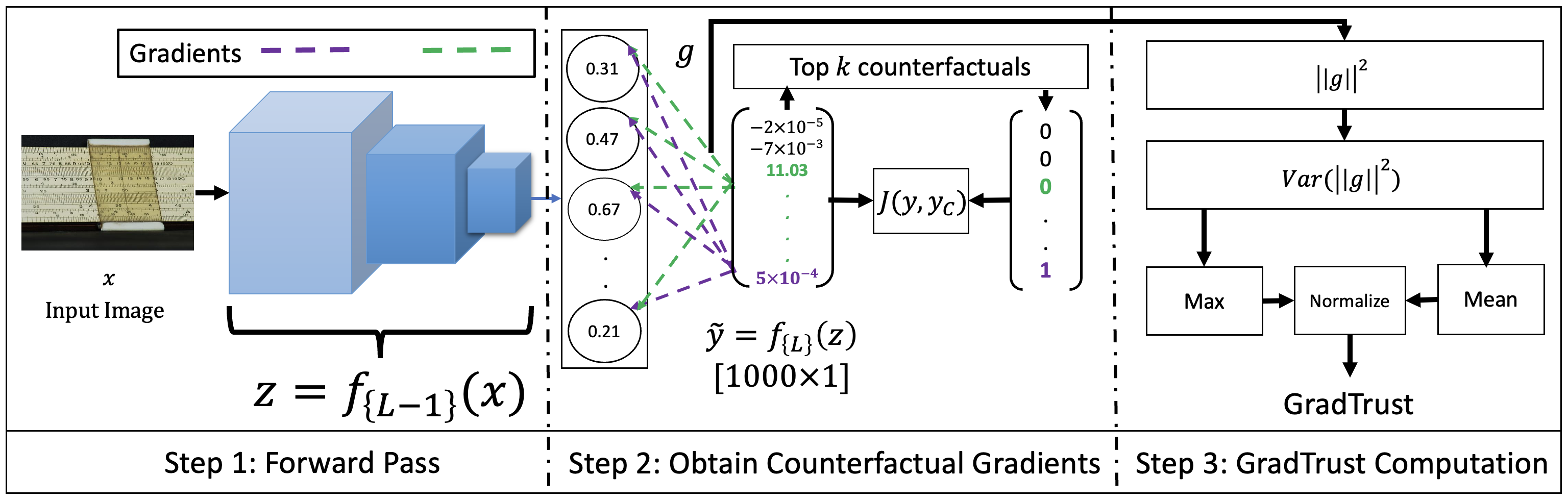}
    \caption{Block diagram of \texttt{GradTrust} quantification.}
    \label{fig:Block}
\end{figure*}

\noindent\textbf{Direct Trust Quantification} The authors in~\cite{jiang2018trust} quantify trust as a function of classification model predictions and a modified nearest neighbor classifier. The modifications involve selectively choosing the right samples from a held-out validation set to quantify trust. In modern neural networks with a large pool of training and validation data, this requirement of obtaining a nearest neighbor classifier is infeasible. Moreover, the distance assumptions made to construct nearest neighbor classifiers are impractical. For instance, trust quantification in~\cite{jiang2018trust} is only negligibly more robust than softmax probability on CIFAR-10 dataset. In this paper, we extract trust quantification from trained neural networks, given a single image, by utilizing counterfactual gradients as features.

\noindent\textbf{Gradients as Features} Gradients provide a measurable change in the network parameters. Recently, a number of works have used gradients as features to characterize data as a function of network weights. This characterization has shown promising results in disparate applications including novelty~\cite{kwon2020novelty}, anomaly~\cite{kwon2020backpropagated}, and adversarial image detection~\cite{lee2022gradient}, image quality assessment~\cite{kwon2019distorted}, , severity detection~\cite{kokilepersaud2022gradient}, and human visual saliency detection~\cite{sun2020implicit}. A number of theories as to their efficacy has been put forward including neurobiological~\cite{prabhushankar2023stochastic}, behavioral~\cite{prabhushankar2022introspective}, and reasoning-based~\cite{prabhushankar2021contrastive}. In this paper, we adopt the interpretation of gradients as encoding the uncertainty of a loss function~\cite{lee2023probing}. 

\section{Methodology}
\label{sec:Method}
\noindent\textbf{Intuition} Qualitatively, we define trust as a region around the existing neural network weights where perturbations in the weight parameters do not change the outputs. Variance of gradients of the neural network weights w.r.t. the loss provides how \textit{spread out} across the classes the required update is to incorporate the loss. At inference and for a well trained network, the true prediction has a minimal loss and backpropagating the prediction leads to minimal variance. However, backpropagating a \textit{counterfactual class} requires larger changes within the network and provides a large variance. Higher this change, larger is the separation between the prediction and the backpropagated counterfactual class and more is the trust in the prediction. This is similar to margin sampling~\cite{balcan2007margin} for the application of active learning where the smallest margin between the prediction and the next best prediction is taken as a measure of uncertainty. By utilizing gradients, we leverage the sensitivity of the learned weights to counterfactuals. Hence, \texttt{GradTrust} is a function of the data, the network weights and the changes in the weights induced by a counterfactual backpropagation. 

\noindent\textbf{Notations} Let $f(\cdot)$ be an $L$ layered neural network trained to distinguish between $N$ classes. If $x$ is any input to the network, the prediction from a classification application is given by $f(x) = y_l$ where $y_l$ is an $(N\times 1)$ vector of logits. The prediction $y$ is the index of the maxima of $y_l$. Consider only the final fully connected layer $f_{L}(\cdot)$ parameterized by weights ($W_L$) and bias ($b_L$). We obtain $y$ as,
\begin{equation}\label{eq:Filter}
\begin{gathered}
    y = W_L^T f_{L-1}(x) + b_L, \\
    \forall  y\in \Re^{N\times 1}, W_L\in \Re^{d_{L-1}\times N}, f_{L-1}(x)\in \Re^{d_{L-1}\times 1},
\end{gathered}
\end{equation}
where $f_{L-1}(x)$ is the flattened output of the network just before the final fully connected layer. The goal of training is construct $f_{L-1}(X)$ to span a linearly separable representation space so that $f_L(\cdot)$ can classify $x \in X$ among $N$ classes via linear projection. In this paper, we provide a trust quantification $r_{x,y}$ for prediction $y$ at inference based on gradients from the weights of the last fully-connected layer $W_L$.

\subsection{GradTrust Quantification}
\label{subsec:GradTrust}
We define \texttt{GradTrust} as the ratio of variance of the maximum counterfactual class gradients to the mean of variances of top-$k$ counterfactual class gradients. Larger the ratio, higher is the region around the weight parameters for predicting some other counterfactual class. We use $k$ likely counterfactual classes in the denominator since for pretrained networks, not all $N$ classes have class probabilities. Likely is defined as the top $k$ logits in the final layer. The block diagram for obtaining \texttt{GradTrust} is provided in Fig.~\ref{fig:Block}. 

The goal is to obtain \texttt{GradTrust} $r_{x,y}$ for a given data point $x$, prediction $y$, and using a loss function $J(\cdot)$. On ImageNet dataset $N = 1000$. Step 1 in Fig.~\ref{fig:Block} involves a forward pass of image $x$ through $f(\cdot)$. The network provides a $1000\times1$ logit vector $y_l = f(x)$. In Step 2 of Fig.~\ref{fig:Block}, the logits are ordered in descending order and the top-$k$ classes are set to ones and all others (including the prediction $y$) are set to zeros to construct the counterfactual vector $y_C$. In this paper, we empirically choose $k = 10$. MSE loss is taken between the logit and counterfactual vectors and backpropagated across the last layer $f_L$. We obtain the gradients $g$ w.r.t. the last layer weight parameters as,
\begin{gather}
    J(y, y_C) = \frac{1}{k-1} \sum_{i=1}^k  (y_i - y_C)^2, \\
    g = \triangledown_{W} J(y, y_C), g \in \mathbb{R}^{d_{(L-1)} \times N}.
\end{gather}
The gradients w.r.t. the weights of the last fully connected layer are extracted as $g$. $g$ is a $(d_{L-1} \times N)$ matrix where $(d_{L-1})$ is the dimensionality of the penultimate layer and $N$ is the number of classes. The $L2-$norm of $g$ is calculated and the variance across each $d_{L-1} \times 1$ weight is taken to obtain a $N\times 1$ vector. Note that each element of this vector stores the \emph{spread} of each counterfactual action. The maximum variance $V[\cot]$ within the vector is extracted. This variance is normalized using the mean of all remaining elements in the vector to obtain \texttt{GradTrust}. Hence, \texttt{GradTrust} is mathematically formulated as,
\begin{equation}
    r_{x,y} = \text{max}\bigg(\frac{V[\lvert\lvert g \rvert\rvert^2]}{\sum_{i = 1, i \neq y}^k V[\lvert\lvert g_i \rvert\rvert^2]}\bigg).
\end{equation}

\begin{table*}[t!]
\footnotesize 
\caption{Results on $50000$ images from ImageNet 2012 validation dataset~\cite{deng2009imagenet}. AUAC and AUFC values are shown for each metric. The top-2 AUAC and AUFC values for every network are bolded. Rows are ordered based on increasing overall accuracy.}
\label{tab:ImageNet}
\setlength{\tabcolsep}{5pt}
\begin{tabularx}{\linewidth}{lccccccccr}
    \toprule
    & \multicolumn{9}{c@{}}{AUAC / AUFC} \\
    \cmidrule(r){2-10}
    Architecture & Softmax & Entropy & NLL & Margin~\cite{balcan2007margin} & ODIN~\cite{liang2017enhancing} & MCD~\cite{gal2016dropout} & GradNorm\cite{huang2021importance} & Purview\cite{lee2023probing} & \textbf{\texttt{GradTrust}} \\
    \midrule
    AlexNet~\cite{krizhevsky2012imagenet} & 72.86/68.43 & 65.02/62.14 & \textbf{83.21}/\textbf{79.37} & 79.04/73.3 & 79.22/75.89 & 54.2/51.59 & 58.85/55.28 & 50.14/48.92 & \textbf{92.09}/\textbf{89.5}\\
    MobileNet~\cite{howard2019searching} & 77.91/74.96 & 71.72/69.9 & \textbf{84.02}/\textbf{81.37} & 83.13/79.1 & 75.95/72.81 & 61.1/59.46 & 70.3/67.28 & 61.85/61.32 & \textbf{93.37}/\textbf{90.58} \\
    ResNet-18~\cite{he2016deep} & 79.01/76.13 & 73.49/71.71 & \textbf{85.38}/\textbf{82.73} & 83.88/79.87 & 81.64/79.26 & 62.91/61.4 & 71.93/69.29 & 64.9/64.01 & \textbf{91.78}/\textbf{88.65}\\
    VGG-11~\cite{simonyan2014very} & 79.95/77.02 & 74.33/72.52 & \textbf{90.55}/\textbf{88.42} & 84.85/80.77 & 85.08/83.33 & 63.19/61.62 & 73.16/70.06 & 65/63.84 & \textbf{91.79}/\textbf{89.18} \\
    ResNet-50~\cite{he2016deep} & 81.63/79.69 & 77.47/76.32 & \textbf{89.23}/\textbf{86.47} & 85.7/82.83 & 84.13/82.21 & 66.35/65.37 & 77.37/75.64 & 71.68/71.01 & \textbf{92.24}/\textbf{90.09} \\
    ResNeXt-32~\cite{xie2017aggregated} & 81.56/79.97 & 78.11/77.15 & \textbf{89.83}/\textbf{87.37} & 85.16/82.81 & 82.77/80.43 & 66.9/66.09 & 78.61/77.28 & 74.06/73.05 & \textbf{91.55}/\textbf{89.18}\\ 
    WideResNet~\cite{zagoruyko2016wide} & 82.25/80.79 & 78.96/78.1 & \textbf{90.84}/\textbf{88.42} & 85.76/83.57 & 84.5/82.26 & 67.72/66.89 & 78.62/77.5 & 74.55/73.85 & \textbf{91.36}/\textbf{89.12} \\
    Efficient-v2~\cite{tan2021efficientnetv2} & \textbf{91.49}/\textbf{87.84} & 80.12/76.69 & 71.44/66.03 & 85.13/81.59 & 54.16/51.53 & 81.8/79.38 & 61.43/57.53 & 77.79/77.48 & \textbf{93.57}/\textbf{89.61} \\
    ConvNeXt-t~\cite{liu2022convnet} & 88.17/86.21 & 85.56/83.88 & 79.19/76.85 & \textbf{90.68}/\textbf{88.26} & 62.51/60.74 & 85.43/83.82 & 70.86/66.25 & 79.16/78.91 & \textbf{89.08}/\textbf{87.23} \\ 
    ResNeXt-64~\cite{xie2017aggregated} & 88.95/84.69 & 85.9/80.71 & \textbf{90.04}/\textbf{87.06} & \textbf{91}/86.62 & 76.61/72.94 & 75.3/70.86 & 73.5/71.64 & 80.2/79.96 & 89.15/\textbf{87.41}\\ 
    Swin-v2-t~\cite{liu2022swin} & 86.05/84.27 & 83.79/82.43 & 86.33/83.14 & \textbf{88.75}/\textbf{86.29} & 79.85/77.09 & 84.64/83.17 & 82.23/80.29 & 77.76/77.39 & \textbf{87.45}/\textbf{85.23}\\
    VIT-b-16~\cite{dosovitskiy2020transformers} & 85.97/84.38 & 84.5/82.9 & 82.94/80.3 & \textbf{88.67}/\textbf{86.5} & 62.74/61,03 & 84.33/82.81 & 78.53/74.6 & 78.02/77.73 & \textbf{87.77}/\textbf{85.85}\\
    Swin-b~\cite{liu2021swin} & 86.18/84.49 & 84.77/83.14 & 79.18/75.52 & \textbf{88.5}/\textbf{86.21} & 68.07/64.59 & 84.69/83.17 & 83.09/81.52 & 80.71/80.45 & \textbf{88.44}/\textbf{86.51} \\
    MaxViT-t~\cite{tu2022maxvit} & 84.08/82.66 & 79.23/78.21 & 80.6/78.85 & \textbf{85.84}/\textbf{84.02} & 47.6/46.27 & 80.07/79.08 & 70.35/68.12 & 80.99/80.7 & \textbf{90.19}/\textbf{88.48} \\
    
\bottomrule
\end{tabularx}\vspace{-2mm}
\end{table*}
\begin{figure}[t]
    \centering
    \includegraphics[width=0.99\linewidth]{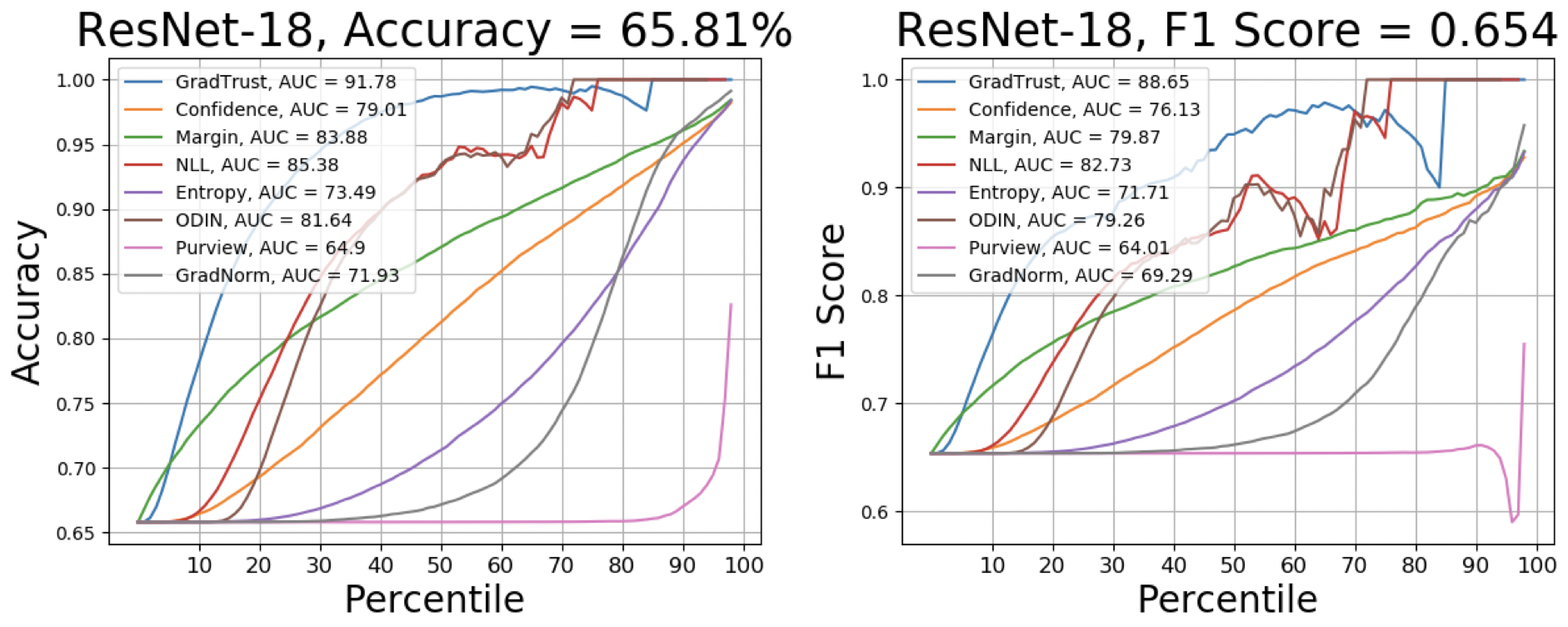}\vspace{-5mm}
    \caption{Accuracy and F1 values of \texttt{GradTrust} and comparison metrics against corresponding percentile bins.}
    \label{fig:AUC}
\end{figure}

\section{Experiments}
\label{sec:Experiments}
In this section, we empirically validate the effectiveness of \texttt{GradTrust}'s ability to detect samples whose classifications are trustworthy. In Section~\ref{subsec:Imagenet}, we show the performance of \texttt{GradTrust} on $14$ CNN and transformer-based architectures.

\subsection{Evaluation}
\label{subsec:Evaluation}
\noindent\textbf{Evaluation strategy} We use the evaluation strategy from~\cite{jiang2018trust}, described below. Given a trained model $f(\cdot)$ and a set of $M$ images, the network provides $M$ individual trust scores for every image. These scores are derived using \texttt{GradTrust} as well as other comparison metrics. For every method, all $M$ scores are binned by their percentile values. We use $100$ bins. The accuracy of the corresponding percentile level is plotted on the \texttt{y-axis} and the percentile level is plotted on the $x-axis$. This is shown for ResNet-18~\cite{he2016deep} architecture in Fig.~\ref{fig:AUC}a. The area under accuracy curve (AUAC) for every method is calculated and displayed. Higher the AUAC, better is the method for detecting relative mispredictions. However, looking at AUAC is by itself insufficient since ImageNet is not a balanced dataset. Hence, we calculate the F1 scores for the same percentile values and similarly plot them in Fig.~\ref{fig:AUC}b. The area under F1 curve (AUFC) is calculated and displayed.

\noindent\textbf{Comparison metrics} We analyze the results of \texttt{GradTrust} against four baselines, two uncertainty estimation methods, and two OOD detection methods. We compare against baseline softmax confidence as well as methods that utilize basic statistics like Entropy, Negative Log Likelihood (NLL), and Margin~\cite{balcan2007margin}. Uncertainty estimation techniques that do not require changes in architectures including MC-Dropout~\cite{gal2016dropout} and gradient-based Purview~\cite{lee2023probing} are analyzed. We notice that both MC-Dropout and Purview provide scores where increase in uncertainty generally relates to a network not trusting its own prediction. Hence, negatives of these two scores are used. Finally, gradient-based out-of-distribution (OOD) techniques including ODIN~\cite{liang2017enhancing} and GradNorm~\cite{huang2021importance} are also compared against. Note that OOD detection techniques typically are evaluated in settings where multiple sources of distributions (generally from different datasets) are considered and require differentiation based on OOD scores. This is different from our setting where scores characterize the trust in predictions on data from the same distribution.

\noindent\textbf{Architectures} In Table~\ref{tab:ImageNet}, we show results on $14$ different architectures, all standard pretrained implementations from PyTorch. Local convolution-based architectures as well as global attention-based transformer models are utilized in the experiments. In all cases the images undergo standard normalization, bilinear interpolation, and $224\times224\times3$ centre cropping. Note that many of the architectures require slightly different mean crop, but for the sake of experimental simplicity, we ignore the specifics. 

\noindent\textbf{Dataset} We use ImageNet~\cite{deng2009imagenet} to validate the performance of \texttt{GradTrust} for image classification. ImageNet is a standard classification benchmark and all PyTorch available models from Table~\ref{tab:ImageNet} are pretrained on ImageNet. In Section~\ref{subsec:Imagenet}, $50000$ validation set images from ImageNet 2012 Recognition Benchmark~\cite{deng2009imagenet} are used to obtain AUAC and AUFC.

\subsection{ImageNet Experiments}
\label{subsec:Imagenet}
The results on ImageNet validation set are shown in Table~\ref{tab:ImageNet}. The rows are arranged in increasing order of classification accuracy. Among $13$ of the considered $14$ architectures, \texttt{GradTrust} is in the top-2 performing methods in both AUAC and AUFC. In ResNeXt-$64\times 4$, NLL and Margin outperform \texttt{GradTrust} in AUAC scores. The current standard of confidence, Softmax, is always outperformed by NLL, Margin, and \texttt{GradTrust}. While the uncertainty-based MCD~\cite{gal2016dropout} and Purview~\cite{lee2023probing} are unimpressive among the initial networks, their results improve with increase in classification accuracy. A better trained network provides a better uncertainty and hence, the uncertainty score matches the misprediction percentile rates. On the other hand, consistent trends do not hold for the OOD detectors ODIN~\cite{liang2017enhancing} and GradNorm~\cite{huang2021importance}. While ODIN performs well in the initial networks, its results fall off among the transformer architectures. Note that OOD scores predict how unfamiliar an image is for a given input and not necessarily if the prediction is trustworthy. Higher accuracy indicates a better familiarity and hence a lower OOD score. These trends however, are not as acute in GradNorm~\cite{huang2021importance}.  

A curious feature is the progressive performance swaps of NLL and Margin classifier along the table. While NLL performs second best among all methods until WideResNet, its results drop significantly towards the end. This is better understood as a function of accuracy. The rows are ordered by increasing accuracy. The NLL loss is more pronounced when the model incorrectly predicts classes. Consider two adjacent data points which are subjected to NLL. The non linear nature of log function can either negligibly change the distance between the two points or dramatically change the distance. Note that the log function penalizes the worse performing data more. However, for networks that perform well, NLL loss for data are low. Hence, when the loss between two data points is low, NLL barely differentiates between the two. In contrast, when the loss of two adjoining data points is high, the log factor dramatically separates the two points. On the other hand, results in the margin classifier increase as we go down the table. As described in the intuition in Section~\ref{sec:Method}, margin classifier is similar to \texttt{GradTrust} in that it uses the current prediction and the next best prediction to calculate the margin score. Newer transformers are more accurate, and hence their margin is reliable compared to older architectures.

\begin{figure*}[t]
    \centering
    \includegraphics[width=1\linewidth]{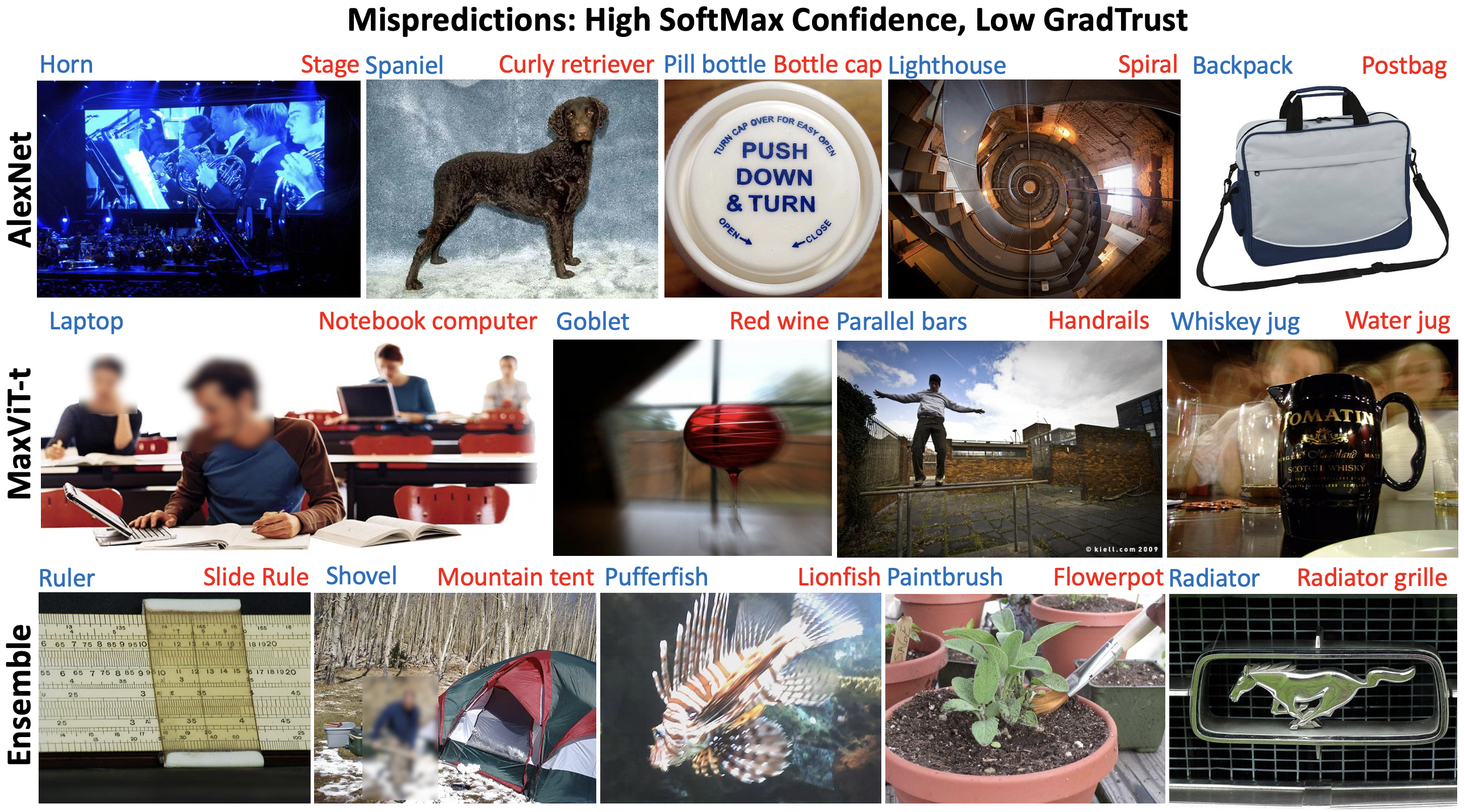}
    \caption{Qualitative analysis of mispredictions on AlexNet (top row), MaxVit-t (middle row) and ensemble mispredictions across all networks from Table~\ref{tab:ImageNet} (bottom row). All displayed images have high softmax and ordered in ascending order of \texttt{GradTrust}.}
    \label{fig:Qualitative}
\end{figure*}

\subsection{Qualitative Analysis}
\label{subsec:Qualitative}
In Fig.~\ref{fig:Qualitative}, we qualitatively showcase the utility of \texttt{GradTrust} on AlexNet~\cite{krizhevsky2012imagenet} (Row 1) and MaxViT-t~\cite{tu2022maxvit} (Row 2) predictions. Specifically, we show mispredictions of both these networks when the softmax confidence is high ($>0.95$ for AlexNet and $>0.9$ for MaxViT-t) and \texttt{GradTrust} is low. In all cases, the ground truths are in blue while the network's mispredictions are in red above each image. The results in Fig.~\ref{fig:Qualitative} illustrate the two kinds of mistrust that networks place in their predictions. On MaxViT-t, \texttt{GradTrust} suggests that the low scores are due to \emph{ambiguity} in resolving between classes. For instance, in the first image of Row 2, the ground truth and prediction class of laptop and notebook computer are ambiguous class labels. Similar statements can be made regarding the other three images. Even for humans, it is hard to differentiate between parallel bars and handrails, and goblet and red wine. In the last image, the contextually surrounding glasses indicate that the label must be a whiskey jug (ground truth) and not a water jug (prediction). On the other hand, the mispredictions from AlexNet in Row 1 indicate that the network's low \texttt{GradTrust} is due to \emph{co-occurring} classes. Both the horn and stage are present in the first image. Similar arguments can be constructed for the pill bottle and bottle cap, and lighthouse and spiral staircase image. On the other hand, the ground truth of backpack seems incorrect for the last image in Row 1. Hence, depending on the network itself, \texttt{GradTrust} detects network mispredictions under class ambiguity and co-occurrence.

In Fig.~\ref{fig:Qualitative} row 3, we show the most frequently mispredicted images from the ImageNet validation set across all the considered networks from Table~\ref{tab:ImageNet} when ordered by \texttt{GradTrust}. In other words, across all $14$ networks from Table~\ref{tab:ImageNet}, the shown images in Fig.~\ref{fig:Qualitative} row 3 have the lowest \texttt{GradTrust}, high softmax, and are misclassified. In the first and third images of row 3, the ground truths of ruler and pufferfish are incorrect. In the shovel and paintbrush labeled images, both the prediction as well as label classes are present. Hence, multiple classes are co-occurring. In the last image of row 3 in Fig.~\ref{fig:Qualitative}, the radiator and radiator grille images are ambiguous. 
\begin{table}[t!]
\footnotesize	
\caption{AUAC and AUFC on $19000$ videos from Kinetics-400 validation  dataset~\cite{kay2017kinetics}. Best results for every network are bolded. Starred $(^*)$ networks use only the first $16$ frames for prediction while unstarred networks that use the first $32$ frames.}
\label{tab:VideoClass}
\setlength{\tabcolsep}{3.2 pt}
\begin{tabularx}{0.99\linewidth}{lcccr}
\toprule
Architecture  & Softmax & NLL & Margin~\cite{balcan2007margin} & \textbf{Proposed}\\
\midrule
ResNet3D-18~\cite{tran2018closer} & 70.65/65.58 & 83.47/79.81 & 76.83/70.49 & \textbf{87.31}/\textbf{83.03}\\
 S3D~\cite{xie2018rethinking} & 62.44/59.56 & 79.24/76.27 & 67.58/63.73 &  \textbf{84.37}/\textbf{80.71}\\
 R2plus1-18~\cite{tran2018closer} & 69.82/66.24 & 82.24/79.87 & 75.34/70.68 & \textbf{87.16}/\textbf{83.88}\\
 MViT-v1$^*$~\cite{fan2021multiscale} & 73.97/70.32 & 69.6/67.02 & \textbf{78.21}/\textbf{73.51} & 76.77/72.46\\
 MViT-v2$^*$~\cite{li2022mvitv2} & 75.06/72.24 & 63.93/61.4 & \textbf{79.15}/\textbf{75.45} & 78.07/74.85\\
\bottomrule
\end{tabularx}
\end{table}

\subsection{Video Classification Experiments}
\label{subsec:Video}
\texttt{GradTrust} is a plug-in framework for any prediction task. In this section, we extend the evaluation setup for video action recognition. Five models from PyTorch library including S3D~\cite{xie2018rethinking}, ResNet3D-18~\cite{tran2018closer}, R2Plus1~\cite{tran2018closer}, MViT-v1~\cite{fan2021multiscale} and MViT-v2~\cite{li2022mvitv2} architectures, pretrained on Kinetics-400 dataset~\cite{kay2017kinetics}, are chosen. Kinetics-400 consists of $400$ actions for a trained model to classify $19000$ validation videos. The results of AUAC and AUFC are shown in Table~\ref{tab:VideoClass}. \texttt{GradTrust} is the top performing metric on three of the five architectures and closely follows Margin in the remaining two.

\subsection{Limitations}
While \texttt{GradTrust} is a better measure for quantifying prediction trust, it's value is unbounded. We use pretrained models for analysis and impose no constraints on the model weights or the resulting logits. This implies that the range of \texttt{GradTrust} differs across networks and data. Other metrics in Table~\ref{tab:ImageNet} including Negative Log Likelihood (NLL), Margin, and popular metrics in literature including Peak Signal-To-Noise Ratio (PSNR) are also unbounded. Using such unbounded measures requires a within-network analysis, as conducted in Table~\ref{tab:ImageNet}. Simple comparison between networks is however, not possible using \texttt{GradTrust} or other unbounded measures. Unlike these techniques, softmax converts any given vector into a probability distribution. Hence, it is bounded between zero and one and will not vary based on architectures. 
\section{Conclusion}
In this paper, we quantify the trust a classification neural network places in its decisions as a function of relative misprediction rates. We leverage the variance in counterfactual gradients to propose \texttt{GradTrust}. The proposed trust measure is a function of the current input, prediction, and the network parameters. \texttt{GradTrust}'s efficacy is showcased on natural images and video data. Across all networks and both data modalities, \texttt{GradTrust} consistently ranks among the top-2 metrics. Surprisingly, simple measures like NLL and Margin are better indicators of misprediction than more state-of-the-art uncertainty and out-of-distribution detection methods. Qualitatively, we show that \texttt{GradTrust} detects images whose ground truth labels are incorrect or ambiguous, or there are co-occurring classes in the image. In all cases, \texttt{GradTrust} builds trust in network predictions.

\bibliographystyle{IEEEbib.bst}
\bibliography{main}

\begin{thebibliography}{10}

\bibitem{toreini2020relationship}
Ehsan Toreini, Mhairi Aitken, Kovila Coopamootoo, Karen Elliott, Carlos~Gonzalez Zelaya, and Aad Van~Moorsel,
\newblock ``The relationship between trust in ai and trustworthy machine learning technologies,''
\newblock in {\em Proceedings of the 2020 conference on fairness, accountability, and transparency}, 2020, pp. 272--283.

\bibitem{varshney2022trustworthy}
Kush~R Varshney,
\newblock ``Trustworthy machine learning,''
\newblock {\em Chappaqua, NY, USA: Independently Published}, 2022.

\bibitem{alregib2022explanatory}
Ghassan AlRegib and Mohit Prabhushankar,
\newblock ``Explanatory paradigms in neural networks: Towards relevant and contextual explanations,''
\newblock {\em IEEE Signal Processing Magazine}, vol. 39, no. 4, pp. 59--72, 2022.

\bibitem{lee2023probing}
Jinsol Lee, Charlie Lehman, Mohit Prabhushankar, and Ghassan AlRegib,
\newblock ``Probing the purview of neural networks via gradient analysis,''
\newblock {\em IEEE Access}, vol. 11, pp. 32716--32732, 2023.

\bibitem{huang2021importance}
Rui Huang, Andrew Geng, and Yixuan Li,
\newblock ``On the importance of gradients for detecting distributional shifts in the wild,''
\newblock {\em Advances in Neural Information Processing Systems}, vol. 34, pp. 677--689, 2021.

\bibitem{lee2022gradient}
Jinsol Lee, Mohit Prabhushankar, and Ghassan AlRegib,
\newblock ``Gradient-based adversarial and out-of-distribution detection,''
\newblock {\em arXiv preprint arXiv:2206.08255}, 2022.

\bibitem{prabhushankar2021extracting}
Mohit Prabhushankar and Ghassan AlRegib,
\newblock ``Extracting causal visual features for limited label classification,''
\newblock in {\em 2021 IEEE International Conference on Image Processing (ICIP)}. IEEE, 2021, pp. 3697--3701.

\bibitem{prabhushankar2022introspective}
Mohit Prabhushankar and Ghassan AlRegib,
\newblock ``Introspective learning: A two-stage approach for inference in neural networks,''
\newblock {\em Advances in Neural Information Processing Systems}, vol. 35, pp. 12126--12140, 2022.

\bibitem{kwon2020backpropagated}
Gukyeong Kwon, Mohit Prabhushankar, Dogancan Temel, and Ghassan AlRegib,
\newblock ``Backpropagated gradient representations for anomaly detection,''
\newblock in {\em European Conference on Computer Vision}. Springer, 2020, pp. 206--226.

\bibitem{abadi2016deep}
Martin Abadi, Andy Chu, Ian Goodfellow, H~Brendan McMahan, Ilya Mironov, Kunal Talwar, and Li~Zhang,
\newblock ``Deep learning with differential privacy,''
\newblock in {\em Proceedings of the 2016 ACM SIGSAC conference on computer and communications security}, 2016, pp. 308--318.

\bibitem{kendall2017uncertainties}
Alex Kendall and Yarin Gal,
\newblock ``What uncertainties do we need in bayesian deep learning for computer vision?,''
\newblock {\em Advances in neural information processing systems}, vol. 30, 2017.

\bibitem{gal2016dropout}
Yarin Gal and Zoubin Ghahramani,
\newblock ``Dropout as a bayesian approximation: Representing model uncertainty in deep learning,''
\newblock in {\em international conference on machine learning}. PMLR, 2016, pp. 1050--1059.

\bibitem{josang2006trust}
Audun Josang, Ross Hayward, and Simon Pope,
\newblock ``Trust network analysis with subjective logic,''
\newblock in {\em Conference Proceedings of the Twenty-Ninth Australasian Computer Science Conference (ACSW 2006)}. Australian Computer Society, 2006, pp. 85--94.

\bibitem{jiang2018trust}
Heinrich Jiang, Been Kim, Melody Guan, and Maya Gupta,
\newblock ``To trust or not to trust a classifier,''
\newblock {\em Advances in neural information processing systems}, vol. 31, 2018.

\bibitem{cheng2020there}
Mingxi Cheng, Shahin Nazarian, and Paul Bogdan,
\newblock ``There is hope after all: Quantifying opinion and trustworthiness in neural networks,''
\newblock {\em Frontiers in artificial intelligence}, vol. 3, pp. 54, 2020.

\bibitem{deng2009imagenet}
Jia Deng, Wei Dong, Richard Socher, Li-Jia Li, Kai Li, and Li~Fei-Fei,
\newblock ``Imagenet: A large-scale hierarchical image database,''
\newblock in {\em 2009 IEEE conference on computer vision and pattern recognition}. Ieee, 2009, pp. 248--255.

\bibitem{he2016deep}
Kaiming He, Xiangyu Zhang, Shaoqing Ren, and Jian Sun,
\newblock ``Deep residual learning for image recognition,''
\newblock in {\em Proceedings of the IEEE conference on computer vision and pattern recognition}, 2016, pp. 770--778.

\bibitem{van2020uncertainty}
Joost Van~Amersfoort, Lewis Smith, Yee~Whye Teh, and Yarin Gal,
\newblock ``Uncertainty estimation using a single deep deterministic neural network,''
\newblock in {\em International conference on machine learning}. PMLR, 2020, pp. 9690--9700.

\bibitem{10053381}
Ryan Benkert, Mohit Prabhushankar, Ghassan AlRegib, Armin Pacharmi, and Enrique Corona,
\newblock ``Gaussian switch sampling: A second order approach to active learning,''
\newblock {\em IEEE Transactions on Artificial Intelligence}, pp. 1--14, 2023.

\bibitem{zhou2022ramifications}
Chen Zhou, Mohit Prabhushankar, and Ghassan AlRegib,
\newblock ``On the ramifications of human label uncertainty,''
\newblock {\em arXiv preprint arXiv:2211.05871}, 2022.

\bibitem{kwon2020novelty}
Gukyeong Kwon, Mohit Prabhushankar, Dogancan Temel, and Ghassan AlRegib,
\newblock ``Novelty detection through model-based characterization of neural networks,''
\newblock in {\em 2020 IEEE International Conference on Image Processing (ICIP)}. IEEE, 2020, pp. 3179--3183.

\bibitem{kwon2019distorted}
Gukyeong Kwon, Mohit Prabhushankar, Dogancan Temel, and Ghassan AlRegib,
\newblock ``Distorted representation space characterization through backpropagated gradients,''
\newblock in {\em 2019 IEEE International Conference on Image Processing (ICIP)}. IEEE, 2019, pp. 2651--2655.

\bibitem{kokilepersaud2022gradient}
Kiran Kokilepersaud, Mohit Prabhushankar, Ghassan AlRegib, Stephanie~Trejo Corona, and Charles Wykoff,
\newblock ``Gradient-based severity labeling for biomarker classification in oct,''
\newblock in {\em 2022 IEEE International Conference on Image Processing (ICIP)}. IEEE, 2022, pp. 3416--3420.

\bibitem{sun2020implicit}
Yutong Sun, Mohit Prabhushankar, and Ghassan AlRegib,
\newblock ``Implicit saliency in deep neural networks,''
\newblock in {\em 2020 IEEE International Conference on Image Processing (ICIP)}. IEEE, 2020, pp. 2915--2919.

\bibitem{prabhushankar2023stochastic}
Mohit Prabhushankar and Ghassan AlRegib,
\newblock ``Stochastic surprisal: An inferential measurement of free energy in neural networks,''
\newblock {\em Frontiers in Neuroscience}, vol. 17, pp. 926418, 2023.

\bibitem{prabhushankar2021contrastive}
Mohit Prabhushankar and Ghassan AlRegib,
\newblock ``Contrastive reasoning in neural networks,''
\newblock {\em arXiv preprint arXiv:2103.12329}, 2021.

\bibitem{balcan2007margin}
Maria-Florina Balcan, Andrei Broder, and Tong Zhang,
\newblock ``Margin based active learning,''
\newblock in {\em International Conference on Computational Learning Theory}. Springer, 2007, pp. 35--50.

\bibitem{liang2017enhancing}
Shiyu Liang, Yixuan Li, and Rayadurgam Srikant,
\newblock ``Enhancing the reliability of out-of-distribution image detection in neural networks,''
\newblock {\em arXiv preprint arXiv:1706.02690}, 2017.

\bibitem{krizhevsky2012imagenet}
Alex Krizhevsky, Ilya Sutskever, and Geoffrey~E Hinton,
\newblock ``Imagenet classification with deep convolutional neural networks,''
\newblock {\em Advances in neural information processing systems}, vol. 25, 2012.

\bibitem{howard2019searching}
Andrew Howard, Mark Sandler, Grace Chu, Liang-Chieh Chen, Bo~Chen, Mingxing Tan, Weijun Wang, Yukun Zhu, Ruoming Pang, Vijay Vasudevan, et~al.,
\newblock ``Searching for mobilenetv3,''
\newblock in {\em Proceedings of the IEEE/CVF international conference on computer vision}, 2019, pp. 1314--1324.

\bibitem{simonyan2014very}
Karen Simonyan and Andrew Zisserman,
\newblock ``Very deep convolutional networks for large-scale image recognition,''
\newblock {\em arXiv preprint arXiv:1409.1556}, 2014.

\bibitem{xie2017aggregated}
Saining Xie, Ross Girshick, Piotr Doll{\'a}r, Zhuowen Tu, and Kaiming He,
\newblock ``Aggregated residual transformations for deep neural networks,''
\newblock in {\em Proceedings of the IEEE conference on computer vision and pattern recognition}, 2017, pp. 1492--1500.

\bibitem{zagoruyko2016wide}
Sergey Zagoruyko and Nikos Komodakis,
\newblock ``Wide residual networks,''
\newblock {\em arXiv preprint arXiv:1605.07146}, 2016.

\bibitem{tan2021efficientnetv2}
Mingxing Tan and Quoc Le,
\newblock ``Efficientnetv2: Smaller models and faster training,''
\newblock in {\em International conference on machine learning}. PMLR, 2021, pp. 10096--10106.

\bibitem{liu2022convnet}
Zhuang Liu, Hanzi Mao, Chao-Yuan Wu, Christoph Feichtenhofer, Trevor Darrell, and Saining Xie,
\newblock ``A convnet for the 2020s,''
\newblock in {\em Proceedings of the IEEE/CVF conference on computer vision and pattern recognition}, 2022, pp. 11976--11986.

\bibitem{liu2022swin}
Ze~Liu, Han Hu, Yutong Lin, Zhuliang Yao, Zhenda Xie, Yixuan Wei, Jia Ning, Yue Cao, Zheng Zhang, Li~Dong, et~al.,
\newblock ``Swin transformer v2: Scaling up capacity and resolution,''
\newblock in {\em Proceedings of the IEEE/CVF conference on computer vision and pattern recognition}, 2022, pp. 12009--12019.

\bibitem{dosovitskiy2020transformers}
A~Dosovitskiy, L~Beyer, A~Kolesnikov, D~Weissenborn, X~Zhai, and T~Unterthiner,
\newblock ``Transformers for image recognition at scale,''
\newblock {\em arXiv preprint arXiv:2010.11929}, 2020.

\bibitem{liu2021swin}
Ze~Liu, Yutong Lin, Yue Cao, Han Hu, Yixuan Wei, Zheng Zhang, Stephen Lin, and Baining Guo,
\newblock ``Swin transformer: Hierarchical vision transformer using shifted windows,''
\newblock in {\em Proceedings of the IEEE/CVF international conference on computer vision}, 2021, pp. 10012--10022.

\bibitem{tu2022maxvit}
Zhengzhong Tu, Hossein Talebi, Han Zhang, Feng Yang, Peyman Milanfar, Alan Bovik, and Yinxiao Li,
\newblock ``Maxvit: Multi-axis vision transformer,''
\newblock in {\em European conference on computer vision}. Springer, 2022, pp. 459--479.

\bibitem{kay2017kinetics}
Will Kay, Joao Carreira, Karen Simonyan, Brian Zhang, Chloe Hillier, Sudheendra Vijayanarasimhan, Fabio Viola, Tim Green, Trevor Back, Paul Natsev, et~al.,
\newblock ``The kinetics human action video dataset,''
\newblock {\em arXiv preprint arXiv:1705.06950}, 2017.

\bibitem{tran2018closer}
Du~Tran, Heng Wang, Lorenzo Torresani, Jamie Ray, Yann LeCun, and Manohar Paluri,
\newblock ``A closer look at spatiotemporal convolutions for action recognition,''
\newblock in {\em Proceedings of the IEEE conference on Computer Vision and Pattern Recognition}, 2018, pp. 6450--6459.

\bibitem{xie2018rethinking}
Saining Xie, Chen Sun, Jonathan Huang, Zhuowen Tu, and Kevin Murphy,
\newblock ``Rethinking spatiotemporal feature learning: Speed-accuracy trade-offs in video classification,''
\newblock in {\em Proceedings of the European conference on computer vision (ECCV)}, 2018, pp. 305--321.

\bibitem{fan2021multiscale}
Haoqi Fan, Bo~Xiong, Karttikeya Mangalam, Yanghao Li, Zhicheng Yan, Jitendra Malik, and Christoph Feichtenhofer,
\newblock ``Multiscale vision transformers,''
\newblock in {\em Proceedings of the IEEE/CVF international conference on computer vision}, 2021, pp. 6824--6835.

\bibitem{li2022mvitv2}
Yanghao Li, Chao-Yuan Wu, Haoqi Fan, Karttikeya Mangalam, Bo~Xiong, Jitendra Malik, and Christoph Feichtenhofer,
\newblock ``Mvitv2: Improved multiscale vision transformers for classification and detection,''
\newblock in {\em Proceedings of the IEEE/CVF Conference on Computer Vision and Pattern Recognition}, 2022, pp. 4804--4814.

\end{thebibliography}

\end{document}